\documentclass[10pt,twocolumn,letterpaper]{article}

\pdfoutput=1   

\usepackage{wacv}
\usepackage{times}
\usepackage{epsfig}
\usepackage{graphicx}
\usepackage{amsmath}
\usepackage{amssymb}
\usepackage{booktabs}
\usepackage[table, dvipsnames, svgnames, x11names]{xcolor}
\usepackage{amsfonts}
\usepackage{textcomp}
\usepackage{marvosym}
\expandafter\let\csname Cross\endcsname\relax
\usepackage{bbding}
\usepackage{array}
\usepackage{lipsum}
\usepackage[normalem]{ulem}   
\usepackage{colortbl}
\usepackage{makecell}
\usepackage{threeparttable}
\usepackage{xfrac}
\usepackage{stfloats}
\usepackage{multirow}

\def\wacvPaperID{0520} 

\wacvfinalcopy 

\ifwacvfinal
\usepackage[colorlinks,
   linkcolor=red,
   citecolor=green
]{hyperref}
\else
\usepackage[pagebackref=true,breaklinks=true,colorlinks,bookmarks=false]{hyperref}
\fi

\begin{document}

\title{Frequency-Aware Self-Supervised Monocular Depth Estimation}

\author{Xingyu Chen\textsuperscript{1}\qquad Thomas H. Li\textsuperscript{1,2,3}\qquad Ruonan Zhang\textsuperscript{1}\qquad Ge Li \Letter \textsuperscript{1}\\
\small
\textsuperscript{1}School of Electronic and Computer Engineering, Peking University
\small
\textsuperscript{2}Advanced Institute of Information Technology, Peking University\\
\small
\textsuperscript{3}Information Technology R\&D Innovation Center of Peking University\\
{\tt\small cxy@stu.pku.edu.cn\qquad tli@aiit.org.cn\qquad zhangrn@stu.pku.edu.cn\qquad geli@ece.pku.edu.cn}\\
\url{https://github.com/xingyuuchen/freq-aware-depth}
}

\maketitle
\thispagestyle{empty}

\begin{abstract}
   We present two versatile methods to generally enhance self-supervised monocular depth estimation (MDE) models.
   The high generalizability of our methods is achieved by solving the fundamental and ubiquitous problems in photometric loss function.
   In particular, from the perspective of spatial frequency, we first propose Ambiguity-Masking to suppress the incorrect supervision under photometric loss at specific object boundaries, the cause of which could be traced to pixel-level ambiguity.
   Second, we present a novel frequency-adaptive Gaussian low-pass filter, designed to robustify the photometric loss in high-frequency regions.
   We are the first to propose blurring images to improve depth estimators with an interpretable analysis.
   Both modules are lightweight, adding no parameters and no need to manually change the network structures.
   Experiments show that our methods provide performance boosts to a large number of existing models, including those who claimed state-of-the-art, while introducing no extra inference computation at all.
\end{abstract}

\mathchardef\mhyphen="2D
\definecolor{aug_our_method_color}{RGB}{187,249,190}
\definecolor{X_color}{RGB}{255,0,0}
\definecolor{checkmark_color}{RGB}{0,176,80}
\definecolor{shallow_gray}{RGB}{220,220,220}

\section{Introduction}\label{sec:intro}

Inferring the depth of each pixel in a single RGB image is a versatile tool for various fields, such as robot navigation~\cite{griffin2020video}, autonomous driving~\cite{wang2019pseudo,you2019pseudo} and augmented reality~\cite{luo2020consistent}.
However, it is extremely difficult to obtain a large number of depth labels from real world, and even expensive Lidar sensors can only obtain depth information of sparse points on the image~\cite{xu2019depth}.
Therefore, a large number of self-supervised MDE researches have been conducted, with accuracy getting closer and closer to supervised methods.
By exploiting the geometry projection constrain, the self-supervision comes from image reconstructions, requiring only known (or estimated) camera poses between different viewpoints.
Though significant progress has been made, there still remains some undiscovered general problems.

\textbf{First.} Many works~\cite{lyu2020hr,jung2021fine,NEURIPS2020_951124d4,yan2021channel,zhu2020edge} concentrated on predicting clearer (sharper) depth of object boundaries.
Despite their success, they mainly relied on well-designed network architectures.
In this work, we show a more fundamental reason for this limitation - from input images.
An interesting observation in Fig.~\ref{fig:boundary}b raises the question: \textit{Does the photometric loss at the object boundaries really indicates inaccurate depth predictions?}
Self-supervised training minimizes the per-pixel photometric loss based on the 2D-3D-2D reprojection~\cite{zhou2017unsupervised,godard2019digging}.
Every single pixel is expected to attach to one deterministic object, otherwise the depth of a \textit{mixed} object is of no physical meaning.
The pixel-level ambiguity (Fig.~\ref{fig:boundary}c), as it happens, manifests as making the object boundary the fused color of two different objects.
These ambiguous pixels belong to no objects in the 2D-3D back-projection (see point cloud in Fig.~\ref{fig:boundary}d), and have no correspondence when evaluating photometric loss (on the target and synthesized images) after 3D-2D reprojection.
As a result, the network always learns irrational loss from them, regardless of its predicted depths.

\textbf{Second.} Intuitively, for a loss function, predictions close to \textit{gt} should have small loss, whereas predictions with large error ought to deserve harsh penalties (loss).
However, photometric loss does not obey this rule in high-freq regions, as shown in Fig.~\ref{fig:illustration_of_pe}.
In such regions, a tiny deviation from \textit{gt} receives a harsh penalty, while a large error probably has an even smaller loss than \textit{gt}.
These \textit{unfairness} comes from high spatial frequency and the breaking of photometric consistency assumption, respectively.
To reduce such \textit{unfairness}, we present a frequency-adaptive Gaussian blur technique called Auto-Blur.
It enlarges the receptive field by \textit{radiating} photometric information of pixels when needed.

To sum up, our contributions are threefold:
\begin{enumerate}
    \item We show the depth network suffers from irrational supervision under the photometric loss at specific boundary areas.
    We trace its cause to pixel-level ambiguity due to the anti-aliasing technique.
    Furthermore, we demonstrate the photometric loss cannot \textit{fairly} and accurately evaluate the depth predictions in high-frequency regions.
    \item To overcome these two problems, we first propose \textbf{Ambiguity-Masking} to exclude the ambiguous pixels producing irrational supervisions.
    Second, we present \textbf{Auto-Blur}, which pioneeringly proves blurring images could universally enhance depth estimators by reducing \textit{unfairness} and enlarging receptive fields.
    \item Our methods are highly versatile and lightweight, providing performance boosts to a large number of existing models, including those claiming SoTA, while introducing no extra inference computation at all.

\end{enumerate}

Despite our superior results, the key motivation of this paper is to shed light on the problems rarely noticed by previous MDE researchers, and wish our analysis and solutions could inspire more subsequent works.

\section{Related Work}\label{sec:related_work}

\subsection{Supervised Depth Estimation}\label{subsec:supervised-depth-estimation}
Plenty recent researches have proved that deep neural networks bring remarkable improvements to MDE models.
Many MDE (or stereo matching~\cite{mayer2016large,Xu_2020_CVPR}) methods are fully supervised, requiring the depth labels collected from RGB-D cameras or Lidar sensors.
Eigen \textit{et al.}~\cite{eigen2015predicting} introduced a multi-scale architecture to learn coarse depth and then refined on another network.
Fu \textit{et al.}~\cite{fu2018deep} changed depth regression to classification of discrete depth values.~\cite{bhat2021adabins} further extended this idea to adaptively adjust depth bins for each input image.
With direct access to depth labels, loss is formulated using the distance between predicted depth and ground truth depth (Scale-Invariant loss~\cite{lee2019big,bhat2021adabins}, $\mathcal{L}_1$ distance~\cite{kendall2017end,Xu_2020_CVPR}), without relying on assumptions such as photometric consistency or static scenes.
~\cite{alhashim2018high,song2021monocular} also computed $\mathcal{L}_1$ loss between the gradient map of predicted and \textit{gt} depth.

\subsection{Self-Supervised Depth Estimation}\label{subsec:self-supervised-depth-estimation}

Self-supervised MDE transforms depth regression into image reconstruction~\cite{garg2016unsupervised,zhou2017unsupervised}.
Monodepth~\cite{godard2017unsupervised} introduced the left-right consistency to alleviate depth map discontinuity.
Monodepth2~\cite{godard2019digging} proposed to use \textit{min.} reprojection loss to deal with occlusions, and auto-masking to alleviate moving objects and static cameras.
In order to produce sharper depth edges,
~\cite{jung2021fine} leveraged the off-the-shelf fine-grained sematic segmentations, ~\cite{yan2021channel} designed an attention-based network to capture detailed textures.
In terms of image gradient, self-supervised methods~\cite{gao2021pdanet,godard2019digging,ramamonjisoa2021single,watson2019self} usually adopt the disparity smoothness loss~\cite{heise2013pm}.
~\cite{kim2018deep} trained an additional `local network' to predict depth gradients of small image patches, and then integrated them with depths from `global network'.
~\cite{li2020enhancing} computed photometric loss on the gradient map to deal with sudden brightness change, but it is not robust to objects with different colors but the same gradients.
Most related to our Auto-Blur is Depth-Hints~\cite{watson2019self}, which helped the network escape from local minima of thin structures, by using the depth proxy labels obtained from SGM stereo matching~\cite{hirschmuller2007stereo}, while we make no use of any additional supervision and are not restricted to stereo datasets.

\section{The Need to Consider Spatial Frequency}\label{sec:the-problems}

This section mainly describes our motivation, specifically, revealing two problems that few previous works noticed.
We begin with a quick review of the universally used photometric loss in self-supervised MDE (Sec.~\ref{subsec:self-supervised-monocular-depth-estimation}), then we demonstrate from two aspects (Sec.~\ref{subsec:problem1} and Sec.~\ref{subsec:problem2}) that the photometric loss function is not a good supervisor for guiding MDE models in some particular pixels or areas.

\subsection{Appearance Based Reprojection Loss}\label{subsec:self-supervised-monocular-depth-estimation}

In self-supervised MDE setting, the network predicts a dense depth image $D_t$ given an input RGB image $I_t$ at test time.
To evaluate $D_t$, based on the geometry projection constraint, we generate the reconstructed image $\tilde{I}_{t+n}$ by sampling from the source images $I_{t+n}$ taken from different viewpoints of the same scene.
The loss is based on the pixel-level appearance distance between $I_t$ and $\tilde{I}_{t+n}$.
Majorities of self-supervised MDE methods~\cite{godard2017unsupervised,ramamonjisoa2021single,godard2019digging,zhou2017unsupervised,lyu2020hr,yang2018deep,luo2019every} adopt the $\mathcal{L}_1 + \mathcal{L}_{ssim}$~\cite{wang2004image} as photometric loss:
\begin{equation}\label{eq:pe}
    \mathcal{L} (I_{t}, \tilde{I}_{t+n}) = \frac{\alpha}{2}(1-SSIM(I_t, \tilde{I}_{t+n})) + (1 - \alpha) \parallel I_{t} - \tilde{I}_{t+n} \parallel_1,
\end{equation}
where $\alpha = 0.85$ by default and $SSIM$~\cite{wang2004image} computes pixel similarity over a $3\times 3$ window.

\subsection{Does Loss at Object Boundary Make Sense?}\label{subsec:problem1}

\begin{figure*}
    \begin{minipage}[b]{0.38\textwidth}
        \centering
        \centerline{\includegraphics[width=0.9\textwidth]{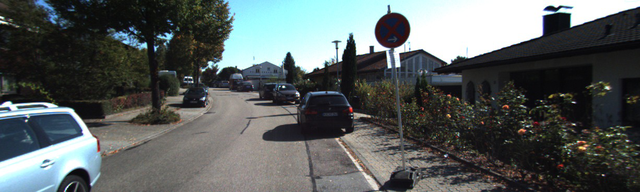}}
        \centerline{(a) Input RGB image}\medskip
        \centerline{\includegraphics[width=0.9\textwidth]{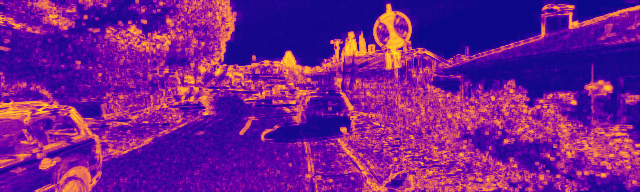}}
        \centerline{(b) Loss map from $8/20^{th}$ epoch}\medskip
    \end{minipage}
    \begin{minipage}[b]{0.3\textwidth}
        \centering
        \centerline{\includegraphics[width=0.95\textwidth]{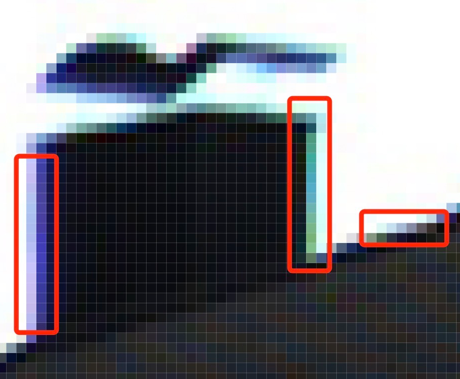}}
        \centerline{(c) Image patch}\medskip
    \end{minipage}
    \begin{minipage}[b]{0.3\textwidth}
        \centering
        \centerline{\includegraphics[width=0.85\textwidth]{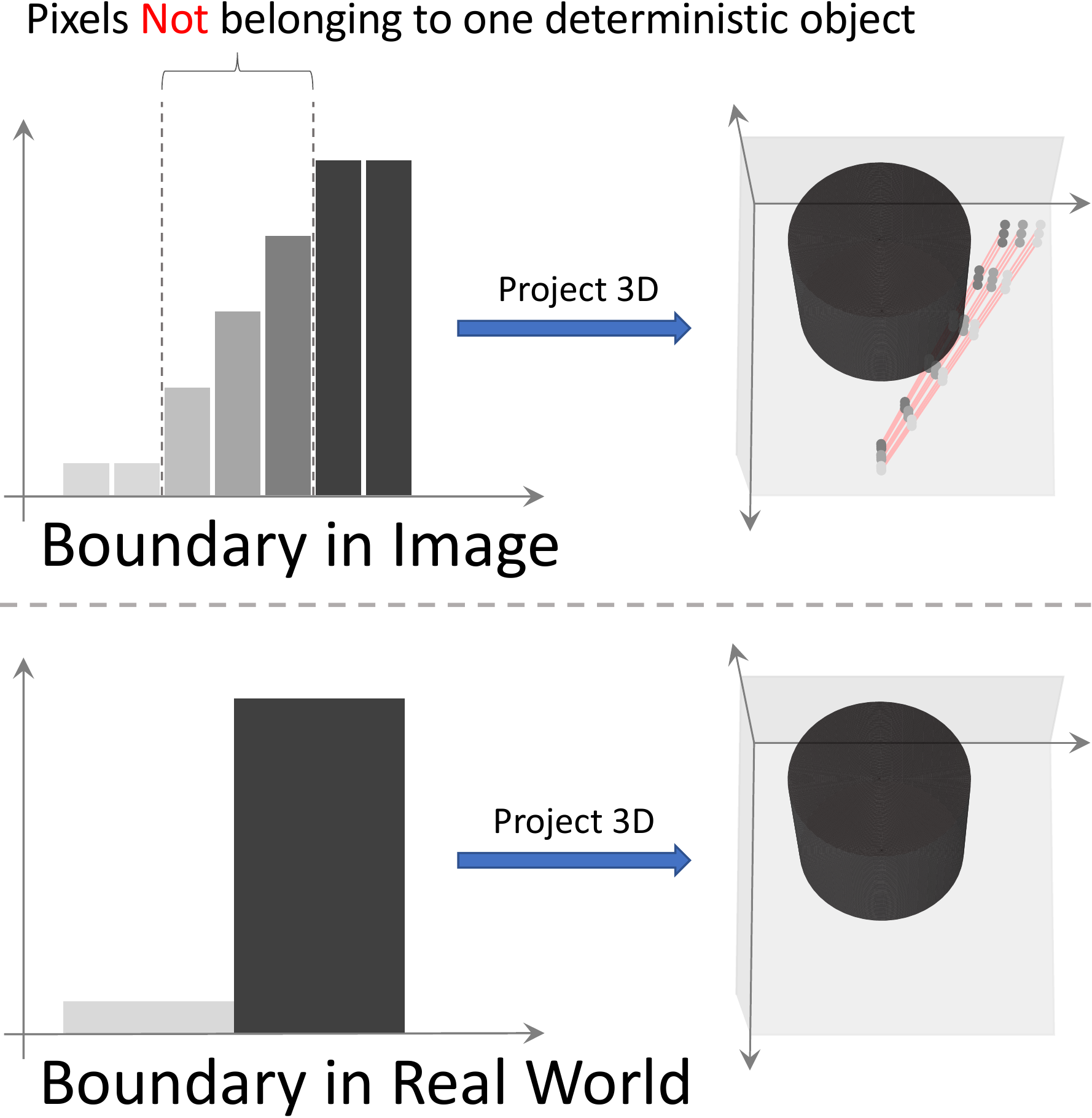}}
        \centerline{(d) Boundary comparison}\medskip
    \end{minipage}
    \caption{
        \textbf{(b)} On most objects, losses appear at object boundaries. \textbf{(c)} The pixels at the boundaries are gradually changed over the junction. However, these colors are ambiguous, \textit{i.e.}, neither from the black chimney nor the white clouds. \textbf{(d)} Object boundaries in the real world are completely mutated, where one single pixel characterizes one deterministic object. However, the ambiguous pixels each contain photometric information for two objects, whereas the network predicts at most one single depth value for them. When projecting the black chimney to 3D point clouds, the ambiguous pixels detach from their main body both spatially and photometrically, regardless of the predicted depths. Hence, no pixels in the synthesized view would match them, resulting in always-large reprojection losses.
    }
    \label{fig:boundary}
\end{figure*}

As seen from Fig.~\ref{fig:boundary}, when training gets to the middle part, the losses appear in two types of regions:
\begin{enumerate}
    \item On the whole object (true-positives).
    Because the estimation of the object's depth (or camera motion) is inaccurate, it reprojects to another object;
    \item At the object boundaries (false-positives).
    Such as the black chimney in the upper right corner.
\end{enumerate}

So why does some loss only appear at the object boundaries, and is it reasonable?
In fact, few works analyzed its cause.
In order to minimize the \textit{per-pixel} reprojection error, the network adjusts \textit{every single pixel's} depth to make it reproject to where it is in the source view.
This process works under the condition that each pixel belongs to \textit{one} deterministic object, since we can never use \textit{one} depth value to characterize a pixel that represents \textit{two} different objects.
However, we illustrate in Fig.~\ref{fig:boundary}c\&d that, the anti-aliasing breaks this training condition by making the object boundary color the weighted sum of both sides' colors.

Specifically, in self-supervised MDE, pixels are first (2D-3D) back-projected to construct the 3D scene using the predicted depths, and then (3D-2D) reprojected to another viewpoint to synthesize the new image.
In the 2D-3D phase, the ambiguous pixels detach from their main body, the 3D points make no physical sense as they do not represent any particular objects, neither spatially nor photometrically (Fig.~\ref{fig:boundary}d).
After the 3D-2D phase, no correspondence from the target image could match these ambiguous colors in the synthesized image, producing large photometric loss.
However, loss should only exist in the area where the depth prediction is incorrect, and these pixels produce unreasonable loss which should not be learnt by the network.

\subsection{Photometric Loss is Unfair in High-Freq Area}\label{subsec:problem2}
\begin{figure*}[t]
    \begin{minipage}{0.4\textwidth}
        \centering
        \centerline{\includegraphics[height=2cm]{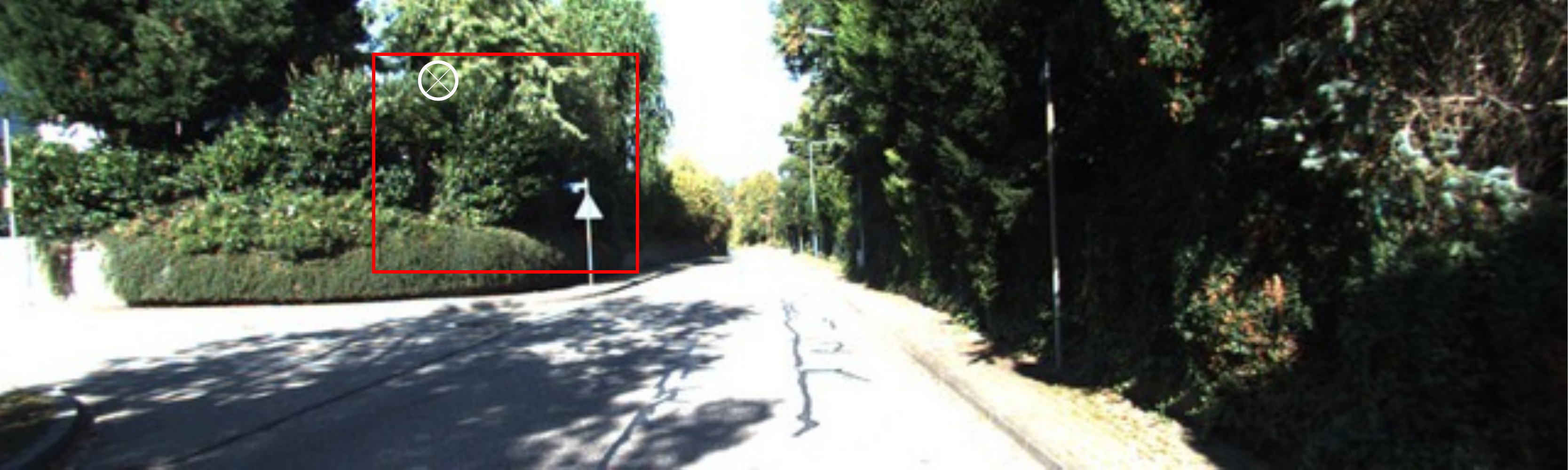}}
        \centerline{Input left image}\medskip
    \end{minipage}
    \hfill
    \begin{minipage}{0.27\textwidth}
        \centering
        \centerline{\includegraphics[height=2cm]{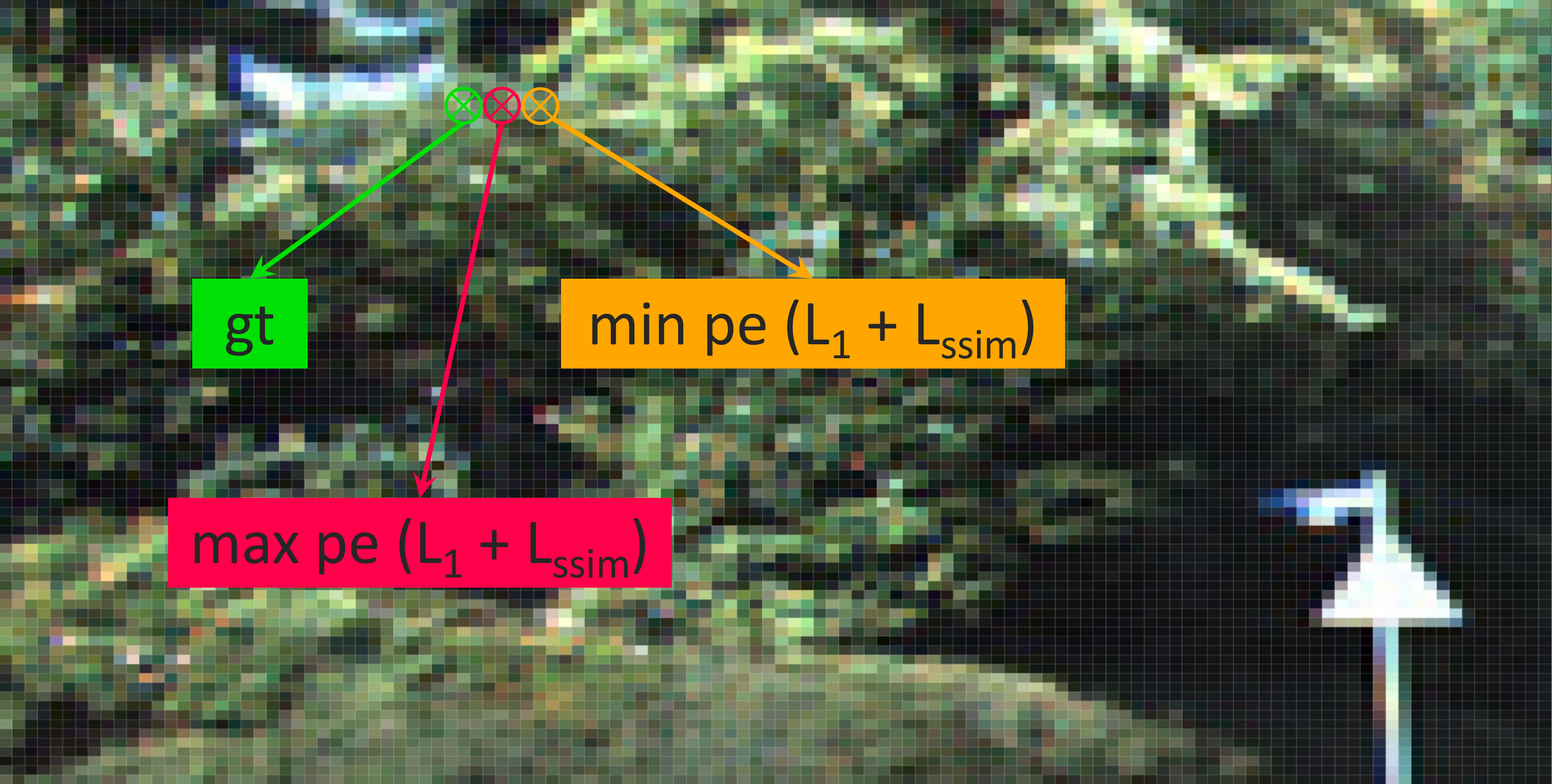}}
        \centerline{Crop of right view}\medskip
    \end{minipage}
    \hfill
    \begin{minipage}{0.27\textwidth}
        \centering
        \centerline{\includegraphics[height=2cm]{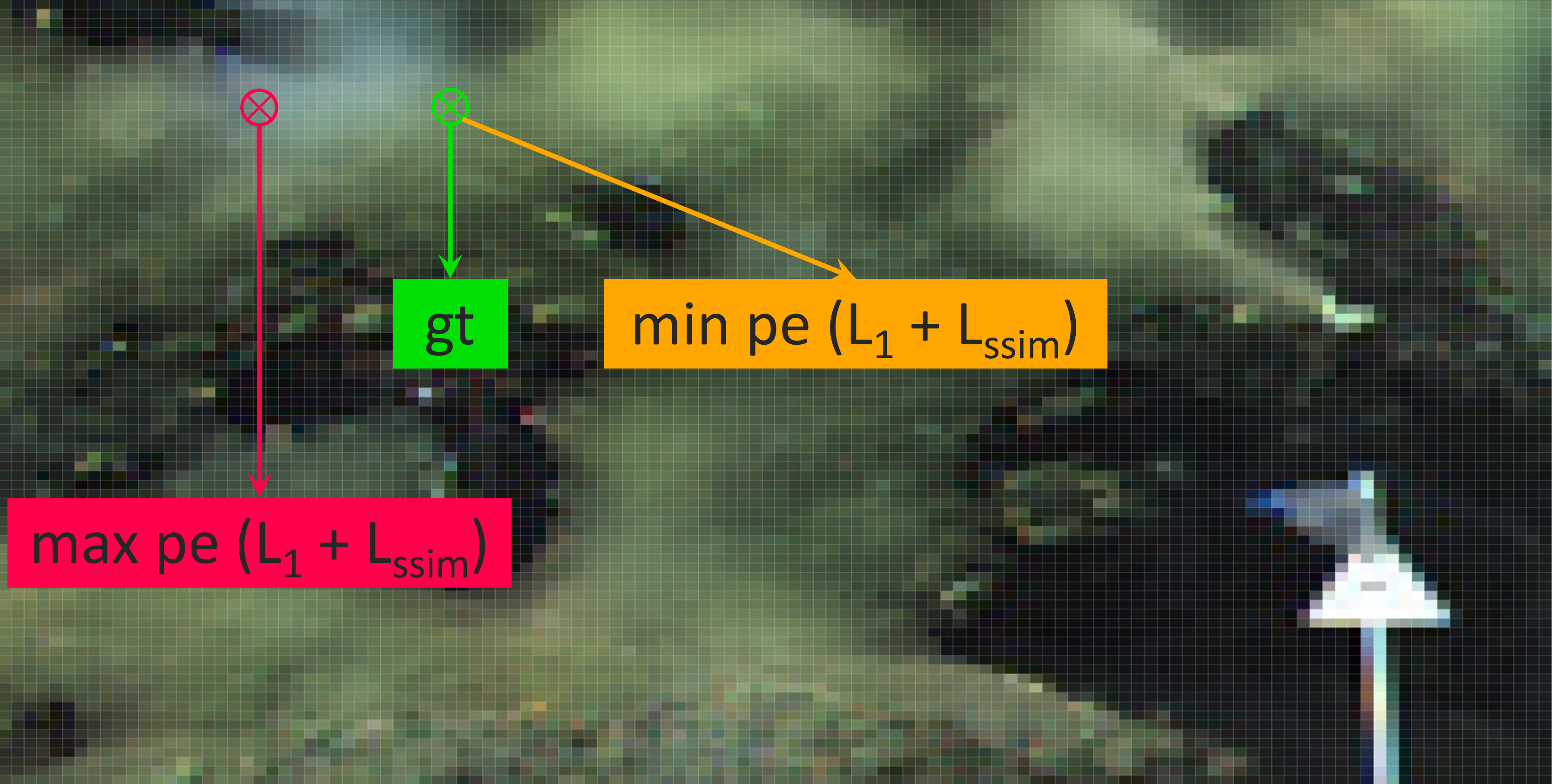}}
        \centerline{With Auto-Blur}\medskip
    \end{minipage}

    \begin{minipage}{0.46\textwidth}
        \centering
        \centerline{\includegraphics[height=3.7cm]{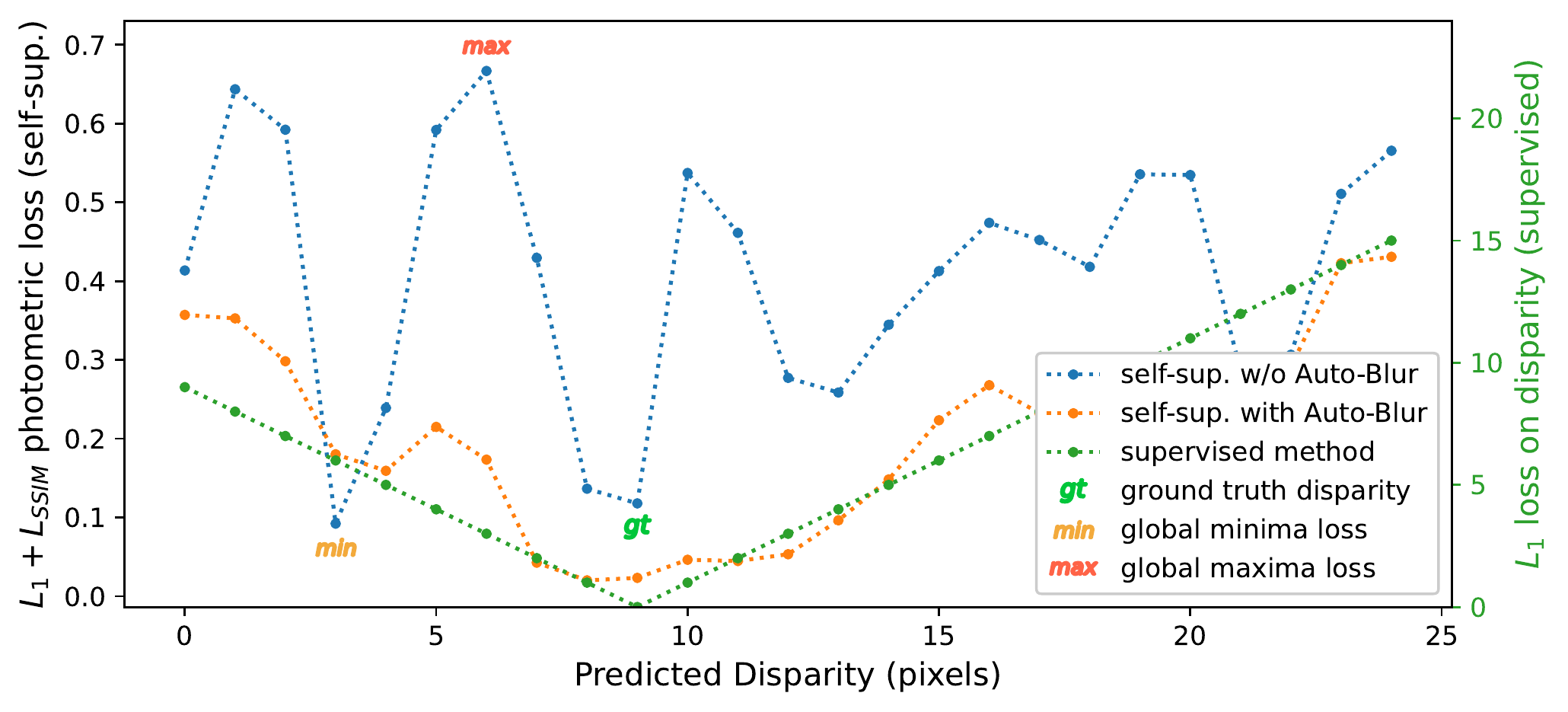}}
    \end{minipage}
    \hfill
    \begin{minipage}{0.31\textwidth}
        \centering
        \centerline{\includegraphics[height=3.6cm]{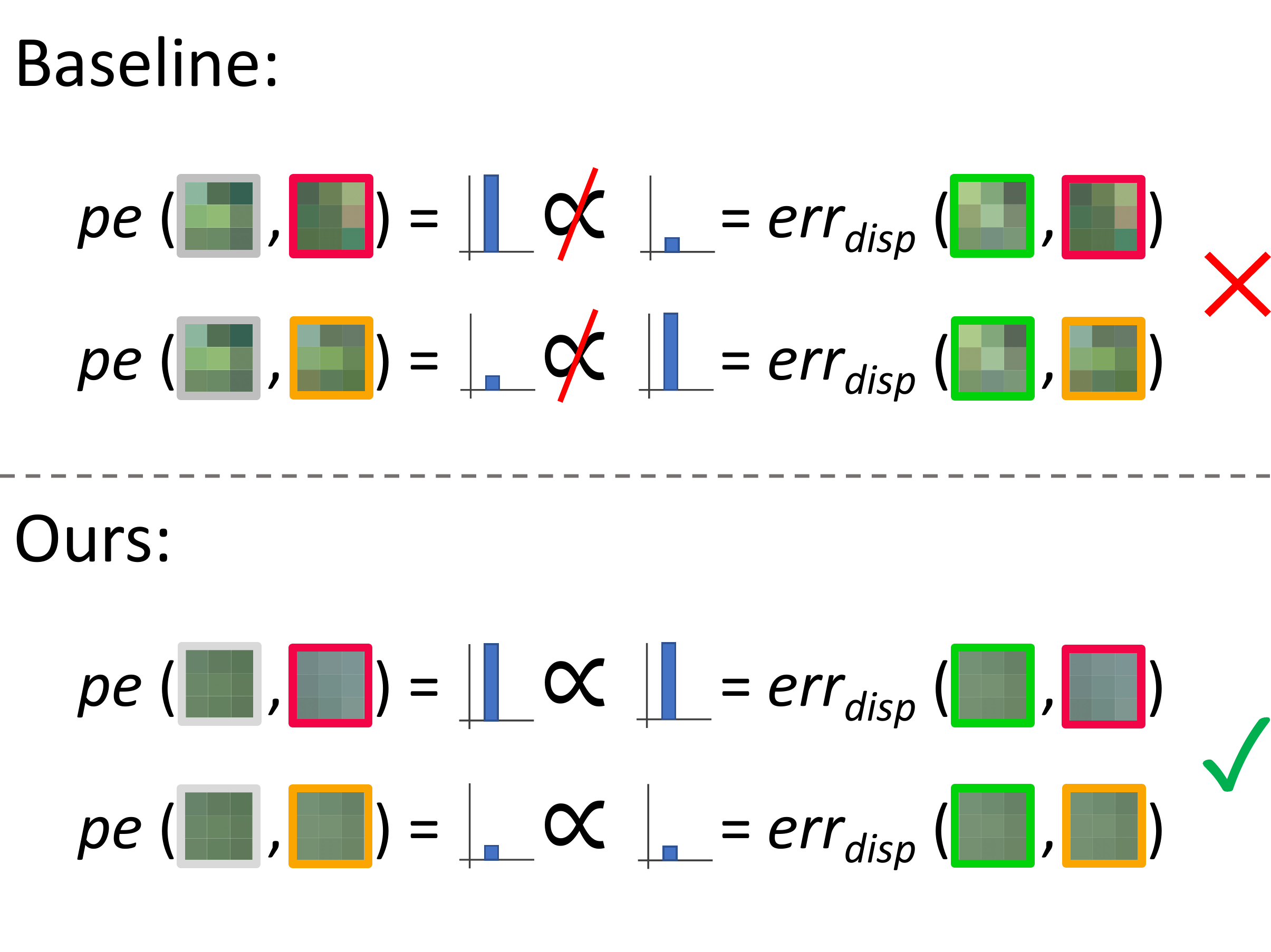}}
    \end{minipage}
    \hfill
    \begin{minipage}{0.21\textwidth}
        \centering
        \centerline{\includegraphics[height=3.8cm]{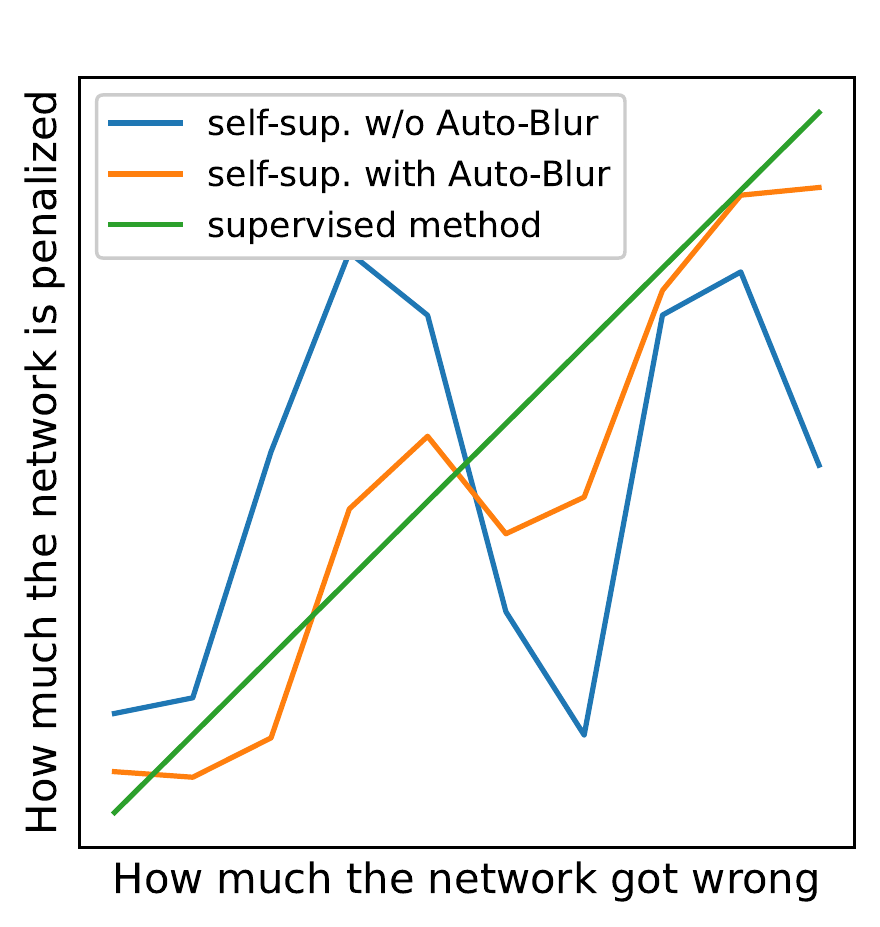}}
    \end{minipage}
    \caption[Caption for LOF]{
        \textbf{Top:} A training image and its crop of the right view (stretched) with and without the proposed Auto-Blur.
        \textbf{Bottom:} Left is the quantitative photometric loss used in self-supervised method \textcolor[RGB]{255,127,14}{with}/\textcolor[RGB]{31,119,180}{without} Auto-Blur and $\mathcal{L}_1$ loss on predicted disparity (Eq.~\ref{eq:supervised_loss}) used in \textcolor[RGB]{44,160,44}{supervised method}. The middle plot ($\propto$: proportional to) shows without Auto-Blur, disparity of \textit{\textcolor[RGB]{234,51,80}{max}} $\mathcal{L}_1 + \mathcal{L}_{ssim}$ photometric loss is instead more accurate than that of \textit{\textcolor[RGB]{243,170,60}{min}} photometric loss; the photometric loss of ground truth is even larger than incorrect disparity, while self-supervised method augmented with Auto-Blur does not suffer from this misjudging. Plot on the right\protect\footnotemark\ is the qualitative analysis of the relationship between network penalty and prediction error. Supervised method exhibits the \textit{absolutely fair} relationship. With Auto-Blur, $\mathcal{L}_1 + \mathcal{L}_{ssim}$ becomes more stable and gets closer to supervised one.
    }
    \label{fig:illustration_of_pe}
\end{figure*}

Before delving into the problem in Fig.~\ref{fig:illustration_of_pe}, we define what is an \textit{absolutely fair loss function} and the \textit{fairness degree}.

\newtheorem{new_definition}{Definition}

\begin{new_definition}[Absolutely Fair Loss Function] \label{def:fair_loss}
    Given a loss function $\mathcal{L}$ for network $\psi$ with ground truth gt, for $\forall x_1, x_2 \in [x_{min}, x_{max}]$, if $|x_1 - gt| < |x_2 - gt|$, then $\mathcal{L}(x_1) < \mathcal{L}(x_2)$.
    We call $\mathcal{L}$ an absolutely fair loss function, with fairness degree = 1 as defined below.
\end{new_definition}

\begin{new_definition}[Fairness Degree] \label{def:fairness_degree}
Given a loss function $\mathcal{L}$ for network $\psi$ with ground truth gt, we compute its fairness degree in the range of $[x_{min},x_{max}]$ subject to:
\begin{equation}\label{eq:fairness_degree}
    \mathcal{D}_{fair} (\mathcal{L}, x_{min}, x_{max}) = \frac{\int_{x_{min}}^{x_{max}} \epsilon\left(\frac{\partial \mathcal{L}(x)}{\partial x}(x-gt) \right) dx} {x_{max}-x_{min}},
\end{equation}
where $\epsilon(\cdot)$ is the indicator function such that $\epsilon(x)=1$ if $x>0$, otherwise $\epsilon(x)=0$.
\end{new_definition}

To better illustrate the problem, we first look at the loss function in one of the \textit{supervised} MDE methods~\cite{kendall2017end}:
\begin{equation}\label{eq:supervised_loss}
\mathcal{L}_{supervised} = \frac{1}{N} \sum_{n=1}^{N} \left \lVert d_n - \hat{d_n} \right \rVert_1,
\end{equation}
where it averages $\mathcal{L}_1$ distances between the predicted depth $d_n$ and ground truth depth $\hat{d_n}$ over all $N$ pixels.
This is an \textit{absolutely fair loss} (Definition~\ref{def:fair_loss}) since the network penalty is (positive) proportional to the prediction error, which always guides the network to converge towards ground truth depth.
In contrast, there are two serious problems in the $\mathcal{L}_1 + \mathcal{L}_{ssim}$ photometric loss as shown in Fig.~\ref{fig:illustration_of_pe}:
\begin{enumerate}
    \item \textit{A small depth estimation error leads to a large loss.} In other words, compared with ground truth, a very slight deviation can produce a large reprojection error, which harshly penalizes the network when its prediction is almost near ground truth;
    \item \textit{A large depth estimation error probably produce an even smaller loss than gt.} Due to the repeated textures in these areas, it is common to mistakenly reproject to another location with the same appearance.
    That is, there are too many local optimums and even false global optimum, interfering with training.
\end{enumerate}

At this point, we could see that being \textit{fair} is a basic requirement for a loss function, otherwise any neural network would be misguided.
Unlike Depth-Hints~\cite{watson2019self} who used additional proxy-labels to help predictions of thin structures escape from local optimum, we focus on improving the \textit{fairness degree} of the loss function itself.

Augmented with the proposed Auto-Blur module, we can achieve a significant improvement in the ill relationship between network penalty and prediction error.
Quantitatively, the \textit{fairness degree}, \textit{i.e.} $\mathcal{D}_{fair}(\mathcal{L}_1 + \mathcal{L}_{ssim}, 0, 15)$, increases from $\frac{8}{15}$ to $\frac{13}{15}$.
Qualitatively, $\mathcal{L}_1 + \mathcal{L}_{ssim}$ no longer suffers from the false global optimum, and looks more like a `V' curve (as the supervised method exhibits on the bottom left of Fig.~\ref{fig:illustration_of_pe}) than before - this indicates a clearer positive proportional relationship, reducing the probability of getting stuck in the local minima.
In the coming section, we show how our Auto-Blur module can relieve this problem without any semantic information.

\footnotetext{Data is from disparity 0$\sim$9 and their losses in the bottom left plot.}

\section{Methodology}
\label{sec:method}

\subsection{Self-supervised Monocular Depth Estimation}\label{subsec:self-supervised-monocular-depth-estimation2}

Following~\cite{zhou2017unsupervised,godard2019digging}, given a monocular and/or stereo video, we first train a depth network $\psi_{depth}$ consuming a single target image $I_t$ as input, and outputs its pixel-aligned depth map $D_t = \psi_{depth}(I_t)$.
Then, except for stereo pairs whose relative camera poses are fixed, we train a pose network $\psi_{pose}$ taking temporally adjacent frames as input, and outputs the relative camera pose $T_{t \rightarrow t+n} = \psi_{pose}(I_t, I_{t+n})$.

Suppose we have access to the camera intrinsics $K$, along with $D_t$ and $T_{t \rightarrow t+n}$, we project $I_t$ into $I_{t+n}$ to compute the sampler \textcolor[RGB]{248,37,68}{$\otimes$}:
\begin{equation}\label{eq:sampler}
    \textcolor[RGB]{248,37,68}{\otimes} = proj \left( D_t,T_{t \rightarrow t+n},K \right).
\end{equation}
The sampler \textcolor[RGB]{248,37,68}{$\otimes$} $\in \mathbb{R}^{H\times W\times 2}$ ($H,W$ represents height and width), which says `for each pixel in $I_t$, where is the corresponding pixel in $I_{t+n}$?'.
We generate the reconstructed image by sampling from $I_{t+n}$ subject to \textcolor[RGB]{248,37,68}{$\otimes$}:
\begin{equation}\label{eq:reconstruct}
    \tilde{I}_{t+n} = \big \langle I_{t+n}, \textcolor[RGB]{248,37,68}{\otimes} \big \rangle,
\end{equation}
where $\langle \cdot,\cdot \rangle$ is the differentiable bilinear sampling operator according to~\cite{godard2019digging}.
The final loss functions consist of photometric loss measured by $\mathcal{L}_1 + \mathcal{L}_{ssim}$ as Eq.~\ref{eq:pe} and edge-aware smoothness loss~\cite{heise2013pm}.

\begin{figure*}
    \centering
    \centerline{\includegraphics[width=0.9\textwidth]{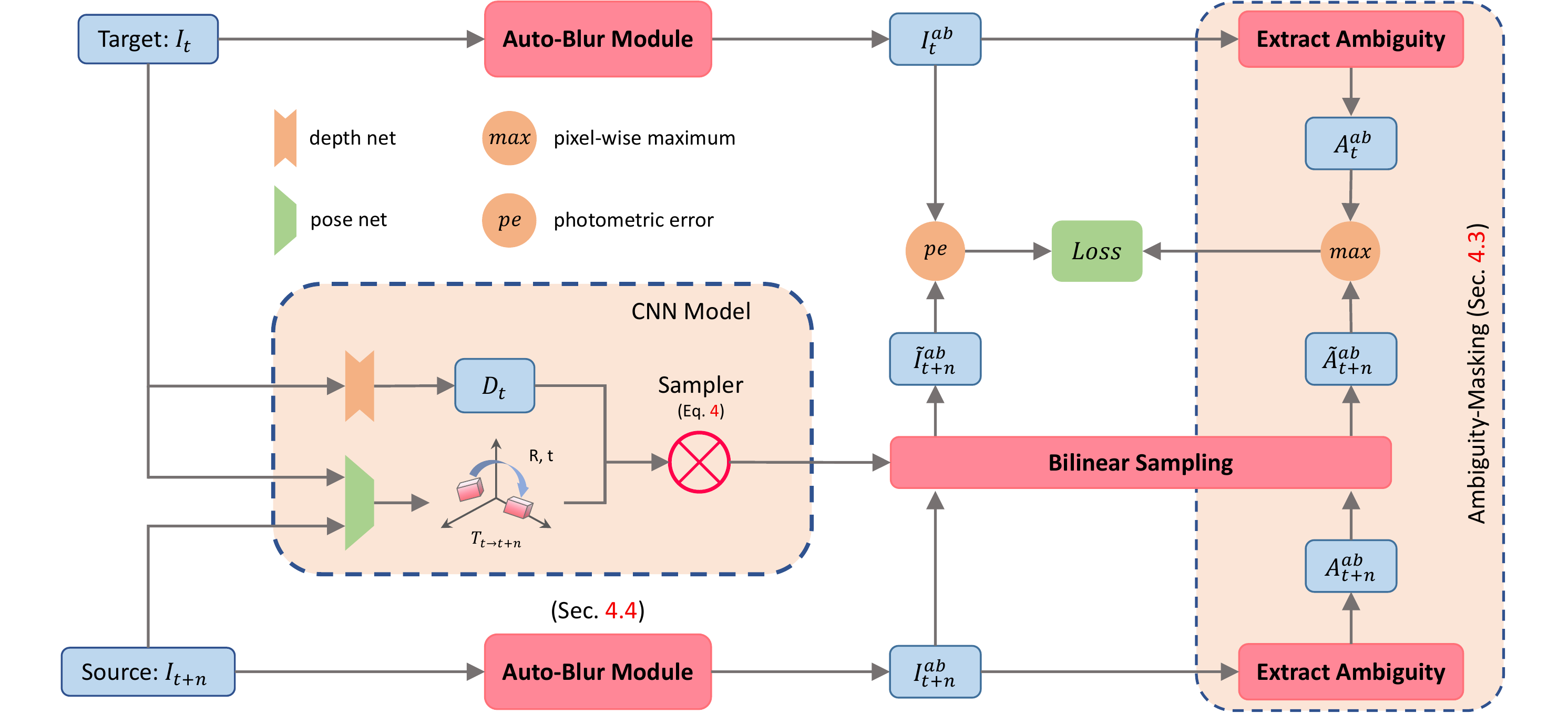}}
    \caption{
        Overview of the proposed method. We propose two approaches to alleviate problems demonstrated in Sec.~\ref{subsec:problem1} and~\ref{subsec:problem2}, respectively, namely \textbf{Ambiguity-Masking} and \textbf{Auto-Blur}. Both are highly versatile, \textit{i.e.} orthogonal to the CNN model architectures.
        The input images are `auto-blurred' adaptively, then input to the photometric loss function to increase its \textit{fairness degree}.
        The Amb.-Masking extracts ambiguities both in the target and reconstructed image, eliminating irrational supervisions.
        Details in Sec.~\ref{subsec:ambiguity-masking} and~\ref{subsec:auto-blur}.
    }
    \label{fig:overview}
\end{figure*}

\subsection{Compute Spatial Frequency}\label{subsec:compute-spatial-frequency}

We first calculate spatial frequencies.
Following~\cite{heise2013pm}, for each pixel (\textit{e.g.}, pixel at $i,j$), we compute differences between its adjacent pixels to represent the gradient.
Specifically, we use $\mathcal{L}_2$ norm of horizontal and vertical differences:
\begin{equation}\label{eq:hori}
    \nabla_{u\pm}(i,j) = I(i,j) - I(i\pm 1,j),
\end{equation}
\begin{equation}\label{eq:vert}
    \nabla_{v\pm}(i,j) = I(i,j) - I(i,j\pm 1),
\end{equation}
\begin{equation}\label{eq:grad_half}
    \nabla_{\pm}(i,j) = \left \lVert \nabla_{u\pm}(i,j), \nabla_{v\pm}(i,j) \right \rVert_2,
\end{equation}
\begin{equation}\label{eq:grad}
    \nabla(i,j) = \left \lVert \frac{\nabla_{u+}(i,j) - \nabla_{u-}(i,j)}{2}, \frac{\nabla_{v+}(i,j) - \nabla_{v-}(i,j)}{2} \right \rVert_2.
\end{equation}

These spatial frequencies allow the following methods to identify their target pixels or regions.
In practice, we adopt $\nabla_{+}$ (Eq.~\ref{eq:grad_half}) in Auto-Blur for simplicity;
while $\nabla$ (Eq.~\ref{eq:grad}) in Ambiguity-Masking for accuracy.

\subsection{Ambiguity-Masking}\label{subsec:ambiguity-masking}

\vspace{0.1cm}
\noindent\textbf{Extract Ambiguity in an Input Image.}
Given an input image $I_t$, we aim to exclude the pixels with ambiguous colors described in Sec.~\ref{subsec:problem1}, \textit{i.e.}, forming the ambiguity map $\mathcal{A}_t$.

The larger the color difference between the adjacent objects, the more ambiguous the pixels located in the object junction.
Hence, we first compute the frequency map $\mathcal{F}_{t}$ as Eq.~\ref{eq:grad}.
Since these pixels are used to smooth the abrupt color changes at object boundary, their colors must be weighted sum of both sides' pixels (see pixels that gradually change from white to black on the sloping roof in Fig.~\ref{fig:boundary}c).
Accordingly, we form a binary mask $\mu$ to pick the high-frequency pixels whose gradients in opposite directions have the opposite sign, either horizontally or vertically, \textit{i.e.},
\begin{equation}\label{eq:same_sign}
    \mu = \left[ \ \nabla_{u+} \cdot \nabla_{u-} < 0 \bigvee \nabla_{v+} \cdot \nabla_{v-} < 0\ \right],
\end{equation}
where $[\cdot]$ is the Iverson bracket.
Then, the initial ambiguity map $\mathcal{A}_t$ for an input $I_t$ is computed as:
\begin{equation}\label{eq:ambiguity_for_image}
    \mathcal{A}_t = \mu \odot \mathcal{F}_t,
\end{equation}
where $\odot$ denotes element-wise multiplication.

\vspace{0.2cm}
\noindent\textbf{Synthesize Ambiguities into a Mask.}
Notably, because photometric loss is based on two images, both target image $I_t$ and reconstructed image $\tilde{I}_{t+n}$ can cause the loss to be untrustworthy.
Thus, we also take $\tilde{I}_{t+n}$ into consideration.

Following~\cite{zhou2017unsupervised,godard2019digging}, for each $I_{t+n}$, we compute the sampler \textcolor[RGB]{248,37,68}{$\otimes_{t+n}$} using $D_{t}$, $T_{t \rightarrow t+n}$ and $K$ subject to Eq.~\ref{eq:sampler}.
Note that \textcolor[RGB]{248,37,68}{$\otimes_{t+n}$} not only contains pixel corresponding relationship used to generate the reconstructed image $\tilde{I}_{t+n}$, but also contains information of how the ambiguities of $I_{t+n}$ affect $\tilde{I}_{t+n}$.
In light of this, we bilinearly sample $\mathcal{A}_{t+n}$ to get which pixels in $\tilde{I}_{t+n}$ are from the ambiguous pixels in $I_{t+n}$ according to \textcolor[RGB]{248,37,68}{$\otimes_{t+n}$}:
\begin{equation}\label{eq:sample_ambiguity}
    \tilde{\mathcal{A}}_{t+n} = \big \langle \mathcal{A}_{t+n}, \textcolor[RGB]{248,37,68}{\otimes_{t+n}} \big \rangle.
\end{equation}

Then, we take the pixel-wise maximum of ambiguities in reconstructed image and target image (intuitively, a \textit{logical or} operation):
\begin{equation}\label{eq:max_ambiguity}
    \mathcal{A}_{t}^{max} = max~\{ \mathcal{A}_{t}, \tilde{\mathcal{A}}_{t+n} \},
\end{equation}
because for each pixel in $\mathcal{L} (I_{t}, \tilde{I}_{t+n})$, the ambiguity from either target image or reconstructed image can both cause its photometric loss to be untrustworthy.
The final ambiguity mask $\mathcal{A}_{t}^{pe}$ is defined by:
\begin{equation}\label{eq:f_func1}
    \mathcal{A}_{t}^{pe} = \left[ \ \mathcal{A}_{t}^{max} < \delta\ \right],
\end{equation}
which is to be element-wise multiplied with $\mathcal{L}(I_{t},\tilde{I}_{t+n})$.
The pseudo-code of the overall algorithm is in Supp.

\subsection{Auto-Blur}\label{subsec:auto-blur}

In order to improve the ill relationship between network penalty and prediction error in high-freq area, we propose Auto-Blur, an adaptive Gaussian low-pass filter in essence.
To be clear, the `auto-blurred' images are only input to the loss function but not to the network, since it is only the photometric loss being unfair and the CNN model expects as much texture as original images to predict accurate depths.

\vspace{0.2cm}
\noindent\textbf{Identify Pixels in High-Frequency Area.}
For simplicity, in Auto-Blur we just adopt $\nabla_{+}$ (Eq.~\ref{eq:grad_half}) as the frequency map $\mathcal{F}_t$ for input image $I_t$.
We first determine whether pixel location $p$ is of high spatial frequency:
\begin{equation}\label{eq:is_hf_pixel}
\mathcal{M}_{is\mhyphen hf\mhyphen pixel} \left( p \right) = \left[\ \mathcal{F}_t \left( p \right) > \lambda \ \right],
\end{equation}
where $\lambda$ is the pre-defined threshold.

Next, we apply average pooling to $\mathcal{M}_{is\mhyphen hf\mhyphen pixel}$ with stride set to $1$, and an Iverson bracket again:
\begin{equation}\label{eq:avg_freq}
\mathcal{M}_{is\mhyphen hf\mhyphen pixel}^{avg} \left( p \right) = \frac{1}{s\times s} \sum_{q \in N_{s\times s} \left( p \right)} \mathcal{M}_{is\mhyphen hf\mhyphen pixel} \left( q \right),
\end{equation}
\begin{equation}\label{eq:is_in_hf_area}
\mathcal{M}_{in\mhyphen hf\mhyphen area} \left( p \right) = \left[\mathcal{M}_{is\mhyphen hf\mhyphen pixel}^{avg} \left( p \right) > \eta\%\right],
\end{equation}
where $s$ is the average pooling kernel size, pixel $q$ belongs to the $s\times s$ neighbors $N_{s\times s}(p)$ of $p$.
Intuitively, if more than $\eta \%$ of the pixels in $N_{s\times s}(p)$ are high-frequency pixels, then $p$ is located in a high-frequency area.

Note that instead of naively averaging $\mathcal{F}_t(q)$ in $N_{s\times s}(p)$ directly, we average the Boolean elements obtained by thresholding $\mathcal{F}_t(q)$ using $\lambda$.
This operation avoids misjudging \textit{thin} object boundaries as high-frequency regions - they are just high-freq pixels themselves, but not in a high-freq \textit{region} that is filled with high-freq pixels.

\begin{table*}[bp]
    \begin{minipage}{0.77\textwidth}
    \scriptsize
        \begin{tabular}{| l | c | c | c || c | c | c | c || c | c | c |}
            \hline
            Method & PP & Data & \begin{tabular}[c]{@{}c@{}}Extra \\ time \end{tabular} & \cellcolor{pink}AbsRel & \cellcolor{pink}SqRel & \cellcolor{pink}RMSE &  \cellcolor{pink}\begin{tabular}[c]{@{}c@{}}RMSE \\ log \end{tabular} & \cellcolor{SkyBlue2}$\delta_1$   & \cellcolor{SkyBlue2}$\delta_2$   & \cellcolor{SkyBlue2}$\delta_3$ \\
            \hline
            \specialrule{0em}{1pt}{1pt}

            \hline Monodepth2 no pt~\cite{godard2019digging} & \textcolor{X_color}{\XSolidBrush} & S & - & 0.130 & 1.144 & 5.485  & 0.232  & 0.831  & 0.932  & 0.968 \\
            \cellcolor{aug_our_method_color}\textbf{+ Ours} & \textcolor{X_color}{\XSolidBrush} & S & \textbf{+ 0ms} & \textbf{0.127} & \textbf{1.086} & \textbf{5.406} & \textbf{0.224} & \textbf{0.832} &  \textbf{0.937} & \textbf{0.971} \\

            \hline
            \specialrule{0em}{1pt}{1pt}
            \hline Monodepth2 M~\cite{godard2019digging} & \textcolor{X_color}{\XSolidBrush} & M & - & 0.115   & 0.903  & 4.863  & 0.193  & 0.877  & 0.959  & 0.981  \\
            \cellcolor{aug_our_method_color}\textbf{+ Ours} & \textcolor{X_color}{\XSolidBrush} & M & \textbf{+ 0ms} & \textbf{0.112}  &  \textbf{0.834}  &  \textbf{4.746}  &  \textbf{0.189}  & \textbf{0.880}  &  \textbf{0.961}  &  \textbf{0.982}  \\

            \hline
            \specialrule{0em}{1pt}{1pt}
            \hline Zhou \textit{et al.}~\cite{zhou2017unsupervised} & \textcolor{X_color}{\XSolidBrush} & M & - & 0.183 & 1.595 & 6.709 & 0.270 & 0.734 & 0.902 & 0.959 \\
            \cellcolor{aug_our_method_color}\textbf{+ Ours}  & \textcolor{X_color}{\XSolidBrush} & M & \textbf{+ 0ms} & \textbf{0.142}  &  \textbf{1.547}  &  \textbf{5.433}  &  \textbf{0.224}  &  \textbf{0.840}  &  \textbf{0.944}  &  \textbf{0.974}  \\

            \hline
            \specialrule{0em}{1pt}{1pt}
            \hline WaveletMonodepth~\cite{ramamonjisoa2021single}  & \textcolor{X_color}{\XSolidBrush} & S & - & 0.109 &  \textbf{0.845} & 4.800  & 0.196 & 0.870 & 0.956 & \textbf{0.980}  \\
            \cellcolor{aug_our_method_color}\textbf{+ Ours} & \textcolor{X_color}{\XSolidBrush} & S & \textbf{+ 0ms} & \textbf{0.108}  &  0.862  &  \textbf{4.786}  &  \textbf{0.194}  &  \textbf{0.875}  &  \textbf{0.957}  &  \textbf{0.980}  \\

            \hline
            \specialrule{0em}{1pt}{1pt}
            \hline Monodepth2 S~\cite{godard2019digging}  & \textcolor{X_color}{\XSolidBrush} & S & - & 0.109 & 0.873  & 4.960  & 0.209  & 0.864  & 0.948  & 0.975  \\
            \cellcolor{aug_our_method_color}\textbf{+ Ours} & \textcolor{X_color}{\XSolidBrush} & S & \textbf{+ 0ms} & \textbf{0.107}  &  \textbf{0.835}  &  \textbf{4.850}  &  \textbf{0.201}  &  \textbf{0.865}  &  \textbf{0.951}  &  \textbf{0.978}  \\

            \hline
            \specialrule{0em}{1pt}{1pt}
            \hline FSRE-Depth~\cite{jung2021fine}  & \textcolor{X_color}{\XSolidBrush} & M & - & \textbf{0.105}  &  0.722  &  4.547  & 0.182 & \textbf{0.886}  & \textbf{0.964}  &  \textbf{0.984}  \\
            \cellcolor{aug_our_method_color}\textbf{+ Ours}  & \textcolor{X_color}{\XSolidBrush} & M & \textbf{+ 0ms} & \textbf{0.105} &  \textbf{0.711}  &  \textbf{4.452}  &  \textbf{0.181}  &  \textbf{0.886}  &  \textbf{0.964}  &  \textbf{0.984}  \\

            \hline
            \specialrule{0em}{1pt}{1pt}
            \hline Monodepth2 MS~\cite{godard2019digging}  & \textcolor{X_color}{\XSolidBrush} & MS & - &\textbf{0.106}  & 0.818  & 4.750  & 0.196 & 0.874 & 0.957 &  0.979  \\
            \cellcolor{aug_our_method_color}\textbf{+ Ours}  & \textcolor{X_color}{\XSolidBrush} & MS & \textbf{+ 0ms} & \textbf{0.106}  &  \textbf{0.797}  &  \textbf{4.672}  &  \textbf{0.187}  &  \textbf{0.887}  &  \textbf{0.961}  &  \textbf{0.982}  \\

            \hline
            \specialrule{0em}{1pt}{1pt}
            \hline CADepth~\cite{yan2021channel}  & \textcolor{X_color}{\XSolidBrush} & S & - & 0.107 & 0.849  &   4.885  &   0.204  &   0.869  &   0.951  &   0.976  \\
            \cellcolor{aug_our_method_color}\textbf{+ Ours}  & \textcolor{X_color}{\XSolidBrush} & S & \textbf{+ 0ms} & \textbf{0.106}  &   \textbf{0.823}  &   \textbf{4.835}  &   \textbf{0.201}  &   \textbf{0.870}  &   \textbf{0.953}  &   \textbf{0.977}  \\

            \hline
            \specialrule{0em}{1pt}{1pt}
            \hline Depth-Hints~\cite{watson2019self}  & \textcolor{X_color}{\XSolidBrush} & S & - & 0.109 &  0.845 & 4.800  & 0.196 & 0.870 & 0.956 & 0.980  \\
            \cellcolor{aug_our_method_color}\textbf{+ Ours}  & \textcolor{X_color}{\XSolidBrush} & S & \textbf{+ 0ms} & \textbf{0.105}  &  \textbf{0.811}  &  \textbf{4.695}  &  \textbf{0.192}  &  \textbf{0.875}  &  \textbf{0.958}  &  \textbf{0.981}  \\

            \hline
        \end{tabular}
    \end{minipage}
    \begin{minipage}{0.22\textwidth}
        \caption{\textbf{Comparison of existing models with and without our methods on KITTI Eigen split~\cite{eigen2014depth}.}
        The \textit{Data} column specifies the training data type: S - stereo images, M - monocular video and MS - stereo video. All models are trained with $192\times 640$ images and Resnet18~\cite{he2016deep} as backbone. All results are not \textbf{P}ost-\textbf{P}rocessed~\cite{godard2017unsupervised}. Metrics are \colorbox{pink}{error metrics \textdownarrow} and \colorbox{SkyBlue2}{accuracy metrics \textuparrow}. Models \colorbox{aug_our_method_color}{augmented with our methods}
        achieve
        better scores
        on almost all metrics, generally.
        No extra inference computation is needed at all.
        }
        \label{tab:kitti_res}
    \end{minipage}
\end{table*}

\vspace{0.2cm}
\noindent\textbf{Blurring Strategy.}
Based on the Gaussian blurred image $I_t^{gb}$ of $I_t$:
\begin{equation}\label{eq:gaussian_blur}
    I_{t}^{gb}(p) = \sum_{q \in N(p)} w^{gb}(q) I_t(q),
\end{equation}
\begin{equation}\label{eq:gaussian_kernel}
    w^{gb}(q) = \frac{1}{\sqrt{2\pi}\sigma} e^{-\frac{\triangle x^2 + \triangle y^2}{2\sigma^2}},
\end{equation}
where pixel $q$ belongs to the neighbors $N(p)$ of $p$, $w^{gb}(q)$ is the weight defined by Gaussian kernel, we compute the final auto-blurred image $I_{t}^{ab}$ subject to:
\begin{equation}\label{eq:auto_blurred}
    I_{t}^{ab} \left( p \right) = w_{blur} \left( p \right) I_t^{gb} \left( p \right) + \left( 1-w_{blur} \left( p \right)  \right) I_{t} \left( p \right),
\end{equation}
\begin{equation}\label{eq:w_blur}
    w_{blur} \left( p \right) = \mathcal{M}_{is\mhyphen hf\mhyphen pixel}^{avg} \left( p \right) \mathcal{M}_{in\mhyphen hf\mhyphen area} \left( p \right),
\end{equation}
where we let $I_{t}^{ab}$ be a weighted sum of $I_{t}^{gb}$ and $I_{t}$, and the more high-frequency pixels around $p$, the more blurry $p$ is.
On the other hand, pixels not in high-frequency areas remain unchanged.

Blurry images are often thought to
degrade the performance of vision systems, instead they benefit the photometric loss, as later analyzed in Sec.~\ref{subsec:analysis-of-auto-blur}.

\section{Experiments}\label{sec:experiments}

\subsection{Implementation Details}\label{subsec:implementation-details}
We set $\delta=0.3$ to extract ambiguous pixels.
We also make use of negative exponential function as an alternative to Eq.~\ref{eq:f_func1}, \textit{i.e.} $\mathcal{A}_{t}^{pe} = e^{-\gamma \mathcal{A}_{t}^{max}}$, where $\gamma=3$.
In Auto-Blur, we set $\lambda=0.2$ to determine whether a pixel is of high frequency, $\eta$ is set to $60$ and $s$ is set to $9$ so that if more than $60\%$ of a pixel's $9 \times 9$ neighbors are of high frequency, then it is regarded as `in high frequency region'.
Our methods support plug and play, so other settings just remain exactly unchanged when embedded into a new baseline, with no more than 10\% additional training time and no extra inference time at all.

\subsection{Quantitative Results}\label{subsec:quantitative-results}
Rather than simply comparing our results to previous SoTA, we run experiments on large numbers of existing models, and compare the results with and w/o our methods in each one of them.
We show our methods lead to superior results, which not only proves that the newly revealed problems are general and ubiquitous, but also makes it possible for future researches to overcome these obstacles.

\vspace{0.1cm}
\noindent\textbf{KITTI} dataset~\cite{geiger2012we} consists of calibrated stereo videos captured from a car driving on the streets in Germany.
The depth evaluation is done on the Lidar point cloud, with all seven of the standard metrics~\cite{eigen2014depth}.
We use the Eigen split of KITTI~\cite{eigen2014depth} and evaluate with Garg’s crop~\cite{garg2016unsupervised}, with standard cap of 80m~\cite{godard2017unsupervised}.
Results are reported in Tab.~\ref{tab:kitti_res}, showing that we help a large number of existing models to achieve better performance.

\begin{table*}
    \scriptsize
    \centering
    \begin{tabular}{ l || c | c | c ||c | c | c | c || c | c | c }
        \Xhline{0.8pt}
        Method & \begin{tabular}[c]{@{}c@{}} Pre- \\ trained \end{tabular} & \begin{tabular}[c]{@{}c@{}} Auto- \\ Blur \end{tabular} & \begin{tabular}[c]{@{}c@{}} Amb.- \\ Masking \end{tabular} & \cellcolor{pink}Abs Rel & \cellcolor{pink}Sq Rel    & \cellcolor{pink}RMSE  & \cellcolor{pink}\begin{tabular}[c]{@{}c@{}}RMSE \\ log \end{tabular} & \cellcolor{SkyBlue2}$\delta\textless 1.25$  & \cellcolor{SkyBlue2}$\delta \textless 1.25^2$   & \cellcolor{SkyBlue2}$\delta \textless 1.25^3$ \\
        \hline

        Base no pt & & & & 0.132 & 1.044 & 5.142 & 0.210 & 0.845 & 0.948 & 0.977  \\
        \arrayrulecolor{shallow_gray}\hline
        Base & \textcolor{checkmark_color}{\checkmark} & & & 0.115   & 0.903  & 4.863  & 0.193  & 0.877  & 0.959  & 0.981  \\

        \arrayrulecolor{shallow_gray}\hline
        Base + Auto-Blur & \textcolor{checkmark_color}{\checkmark}  & \textcolor{checkmark_color}{\checkmark}  & & 0.113  &  0.858  &  4.837  &  0.192  &  0.876  &  0.959  &  0.981  \\

        \arrayrulecolor{shallow_gray}\hline
        Base + Amb.-Mask & \textcolor{checkmark_color}{\checkmark}  & & \textcolor{checkmark_color}{\checkmark} & 0.113 & 0.871 & 4.785 &  0.191  &  \textbf{0.880}  &  0.960 & \textbf{0.982}  \\
        \arrayrulecolor{shallow_gray}\hline
        \textbf{Our full model} & \textcolor{checkmark_color}{\checkmark}  & \textcolor{checkmark_color}{\checkmark} & \textcolor{checkmark_color}{\checkmark} & \textbf{0.112} & \textbf{0.834} & \textbf{4.746} & \textbf{0.189} & \textbf{0.880}  &  \textbf{0.961}  &  \textbf{0.982}  \\
        \Xhline{0.8pt}
    \end{tabular}
    \vspace{0.2cm}
    \caption{\textbf{Ablation on Eigen split~\cite{eigen2014depth}.} The baseline model with none of our contributions and without ImageNet~\cite{deng2009imagenet} pretraining performs poorly. With any of our two approaches, the performance gets improved, and our full model performs the best. All models are trained with $192\times 640$ monocular videos and Resnet18~\cite{he2016deep} as backbone, with results not post-processed~\cite{godard2017unsupervised}.}
    \label{tab:abalation}
\end{table*}

\vspace{0.1cm}
\noindent\textbf{Other Datasets.} To fully justify our benefits, we also carried out experiments on CityScapes~\cite{cordts2016cityscapes} and NYUv2~\cite{silberman2012indoor}.
Again, our methods consistently bring significant improvements to the existing model.
Note that CityScapes even witnesses more significant improvements than KITTI.
In order to further validate our generalizability, we also report results of training on KITTI but evaluating on CS and NYUv2.
\begin{table}[hb]
    \scriptsize
    \centering
    \begin{threeparttable}
        \begin{tabular}{| l | c | c | c | c | c | c |}
            \hline
            Method & Train / Test & \cellcolor{pink}AbsR & \cellcolor{pink}SqR & \cellcolor{pink}log$_{10}$ & \cellcolor{pink}RMSE
            & \cellcolor{SkyBlue2}$\delta_1$
            \\
            \hline
            \specialrule{0em}{1pt}{1pt}
            \hline

            MD2~\cite{godard2019digging} & \multirow{2}{*}{CS / CS} & 0.129 & 1.569 & -- & 6.876 &
            0.849
            \\
            \cellcolor{aug_our_method_color}\textbf{+ Ours}\tnote{$*$} & & \textbf{0.125} & \textbf{1.356} & -- & \textbf{6.618} &
            \textbf{0.856}
            \\
            \hline
            \specialrule{0em}{1pt}{1pt}
            \hline

            MD2~\cite{godard2019digging} & \multirow{2}{*}{KITTI / CS} & 0.163  & 1.883 & -- & 8.967 &
            0.757
            \\
            \cellcolor{aug_our_method_color}\textbf{+ Ours} & & \textbf{0.160} & \textbf{1.854} & -- &  \textbf{8.954} &
            \textbf{0.764}
            \\
            \hline
            \specialrule{0em}{1pt}{1pt}
            \hline

            MD2~\cite{godard2019digging} & \multirow{2}{*}{KITTI / NYUv2} & 0.399 & 0.679 & 0.159 & 1.227 &
            0.420
            \\
            \cellcolor{aug_our_method_color}\textbf{+ Ours} & & \textbf{0.370} & \textbf{0.610} & \textbf{0.142} & \textbf{1.133} &
            \textbf{0.459}
            \\
            \hline
        \end{tabular}
        \begin{tablenotes}
            \scriptsize
            \item[$*$] ~Only $\delta$ needs to be fine-tuned to 0.4 in CityScapes to reach the best performance.
        \end{tablenotes}
    \end{threeparttable}
    \vspace{0.1cm}
    \caption{\textbf{Generalization to other datasets.} Images size: NYUv2 - $256\times 320$; \textbf{C}ity\textbf{S}capes - $416\times 128$ (with preprocessing from~\cite{zhou2017unsupervised}). One more metric $log_{10}$ is reported in NYUv2.}
    \label{tab:other_dataset}
\end{table}

\subsection{Ablation Study}\label{subsec:ablation-study}

We validate our components in Tab.~\ref{tab:abalation} and then analyse our design decisions in detail.

\textit{\textbf{Can the loss function average out the irrational loss?}}
Obviously, the proportion of the ambiguous pixels is low compared to the whole image.
So, does our Amb.-Masking technique really matter?
We made statistics for the ambiguous pixels over $10^3$ batches and report the \textit{mean} value of each metric in Tab.~\ref{tab:percentage}.
Although the number percentage of the ambiguous pixels is not high, the key point is that they each have large irrational loss.
As a result, $\sim$20\% of the final photometric loss comes from these unreasonable pixels, which is actually not a low proportion (almost doubles from the number proportion).
\begin{table}[hb]
    \scriptsize
    \centering
    \begin{tabular}{l *3{c}}
        \Xhline{0.8pt}
        \specialrule{0em}{0.5pt}{0.5pt}
        ~ & Number $\%$ & Photometric Loss & Loss Value $\%$ \\
        \specialrule{0em}{0.5pt}{0.5pt}
        \hline
        \specialrule{0em}{0.5pt}{0.5pt}
        Ambiguous pixels & $10.66\%$ & 0.2415 & $19.43\%$ \\
        Other pixels & $89.34\%$ & 0.1195 & $80.57\%$ \\
        \specialrule{0em}{0.5pt}{0.5pt}
        \Xhline{0.8pt}
    \end{tabular}
    \vspace{0.2cm}
    \caption{Statistical mean values of the ambiguous pixels obtained from $10^3$ training batches.}
    \label{tab:percentage}
\end{table}

\textit{\textbf{How Auto-Blur cooperate with image pyramid loss?}}
Previous works~\cite{godard2019digging,jung2021fine,watson2019self} adopt an image pyramid to evaluate the photometric loss, where a low resolution image is similar to smoothing input images as ours.
So, based on image pyramid loss, how can our Auto-Blur still help?
(\textit{\romannumeral1}) We are texture-specific.
We adaptively (Eq.~\ref{eq:auto_blurred}\&\ref{eq:w_blur}) smooth the high-freq regions which could confuse the photometric loss and keep the original low-freq regions, whereas the pyramid roughly smooths the whole image, which further weakens the supervision signal in texture-less regions that is already weak.
(\textit{\romannumeral2}) We enlarge the receptive field by making each pixel attach the photometric information of its surroundings, thus no information loses.
While downsampling, \textit{e.g.}, could directly erase a two-pixel wide pole.

\begin{figure}
    \centering
    \centerline{\includegraphics[width=\columnwidth]{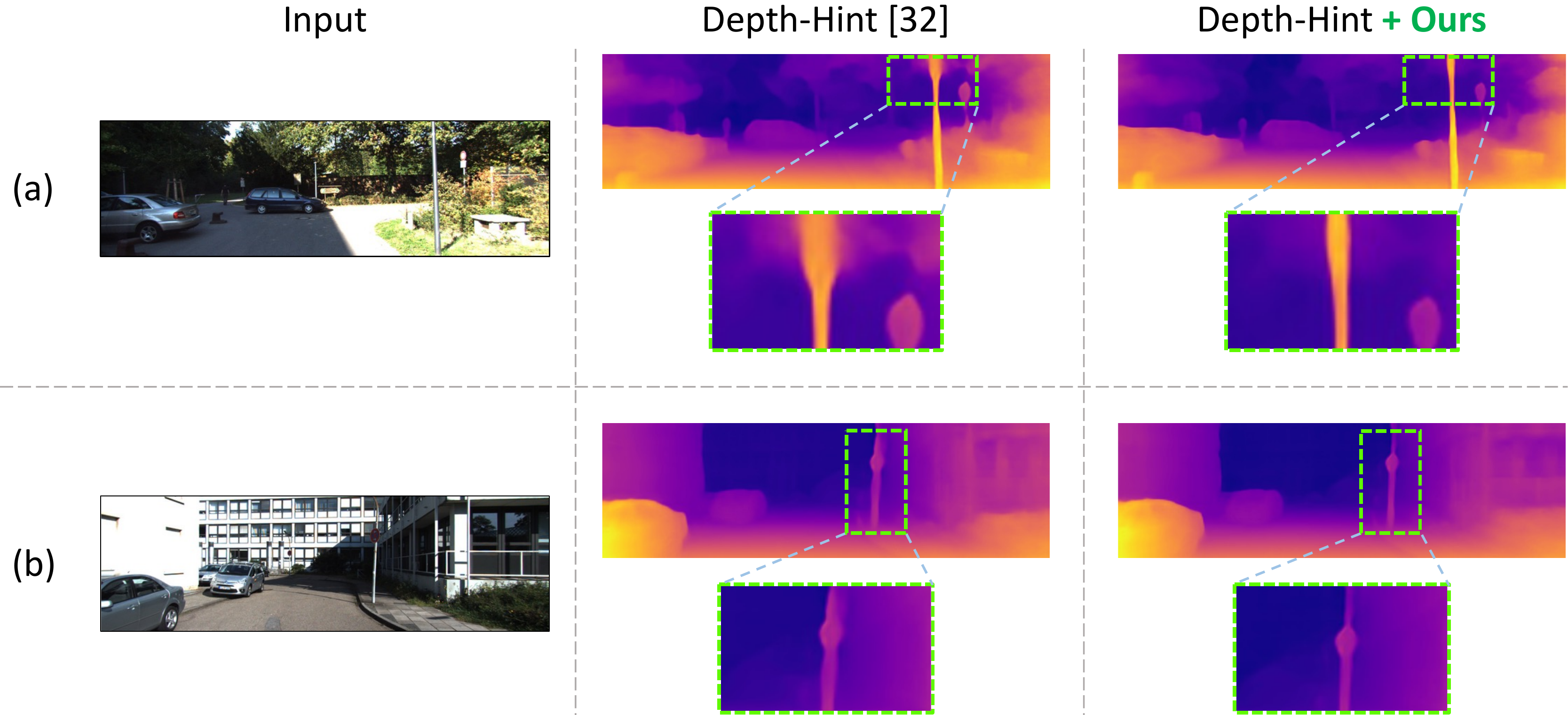}}
    \caption{
        \textbf{Qualitative comparisons}. \textbf{(a)} Auto-Blur enlarges receptive fields (Sec.~\ref{subsec:analysis-of-auto-blur}), helping the pole distinguish from the high-freq background. \textbf{(b)} Amb.-Mask helps to exclude ambiguous pixels whose depths are neither from the pole nor the building.
    }
    \label{fig:visualization}
\end{figure}

\textit{\textbf{Hyper-params decisions.}} We study the hyper-params in Tab.~\ref{tab:abalation_hyperparams}.
For threshold $\lambda$, $\downarrow\!\lambda$ would wrongly smooth the texture-less regions, as the already-weak supervision signal on them will be further weakened.
$\uparrow\!\lambda$ would miss some pixels in high-freq regions which could confuse the photometric loss as illustrated in Fig.~\ref{fig:illustration_of_pe}.
For kernel size $s$ in Auto-Blur, if $\downarrow\!s$, the receptive field could not be effectively enlarged when measuring pixel similarity.
If $\uparrow\!s$, the central pixel's contribution (its own characteristic color) is reduced since the Gaussian distribution gets `shorter' and `wider'.
See Supp. for ablations of all hyper-params.
\begin{table}[h]
    \scriptsize
    \begin{tabular}{c| *3{|c} |c c| *3{|c}}
        \Xhline{0.8pt}
        $\lambda$ & \cellcolor{pink}AbsR & \cellcolor{pink}SqR &  \cellcolor{SkyBlue2}$\delta_1$ & & $s$ & \cellcolor{pink}AbsR & \cellcolor{pink}SqR &  \cellcolor{SkyBlue2}$\delta_1$ \\
        \hline
        \specialrule{0em}{0.5pt}{0.5pt}
        0.15 & 0.113 & 0.844 & 0.879 && 7 & \textbf{0.112} & 0.836  & 0.878  \\
        \arrayrulecolor{shallow_gray}\hline
        0.20 & \textbf{0.112} & \textbf{0.834} & \textbf{0.880} && 9  & \textbf{0.112} & \textbf{0.834}  &  \textbf{0.880}  \\
        \arrayrulecolor{shallow_gray}\hline
        0.25 & 0.113 & 0.881 & 0.877 && 11 & 0.113 & 0.868  &  0.877  \\

        \specialrule{0em}{0.5pt}{0.5pt}
        \Xhline{0.8pt}
    \end{tabular}
    \vspace{0.12cm}
    \centering
    \caption{\textbf{Ablations on hyper-params.} Full results in Supp.}
    \label{tab:abalation_hyperparams}
\end{table}

\subsection{Interpretable Analysis of Auto-Blur}\label{subsec:analysis-of-auto-blur}

\definecolor{mismatch_color}{RGB}{248,37,24}
\definecolor{gt_color}{RGB}{39,165,77}
\begin{figure*}
    \centering
    \centerline{\includegraphics[width=0.9\textwidth]{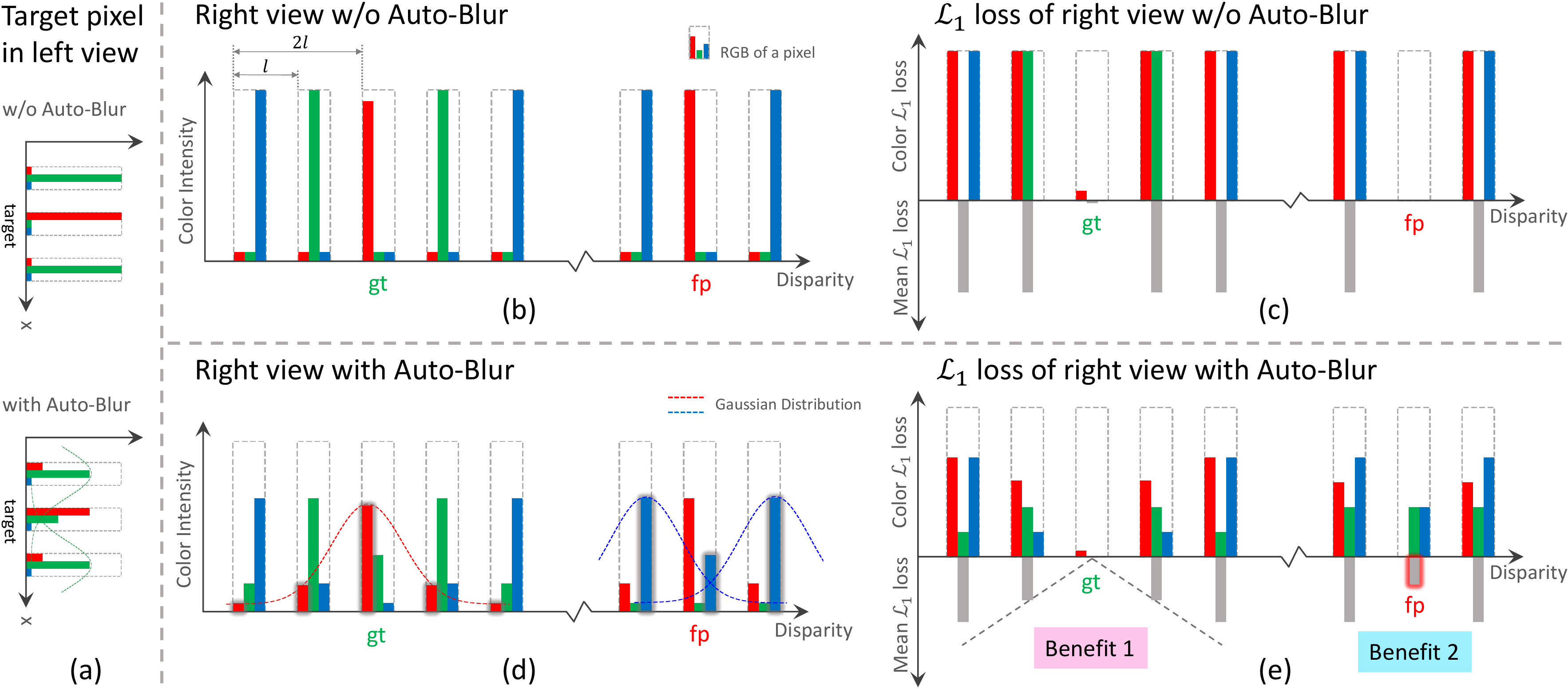}}
    \caption{
        \textbf{An example of why Auto-Blur works.} \textbf{(a)} The color intensity of the target pixel (and its neighbours) in left view with and without Auto-Blur. \textbf{(b)} Original right view. The red \textit{gt} pixel is located in a high-frequency area filled with $R,G,B$ pixels, and there is a target-like (red) pixel in the distance, \textit{i.e} the \textcolor{mismatch_color}{\textit{f}}\textit{alse}-\textcolor{mismatch_color}{\textit{p}}\textit{ositive}. \textbf{(c)} Losses for all estimated disparities are the same except \textcolor{gt_color}{\textit{gt}} and \textcolor{mismatch_color}{\textit{fp}}. Notably, \textcolor{mismatch_color}{\textit{fp}} shows a smaller loss than \textcolor{gt_color}{\textit{gt}}. \textbf{(d)} Under the action of Gaussian kernel, all pixels affect the surroundings and are affected by surroundings. Note that all color channels follow Gaussian distribution independently, the red and blue are highlighted just to illustrate benefit 1 and 2. \textbf{(e)} In contrast to (c), \textbf{Benefit 1:} The augmented $\mathcal{L}_1$ loss becomes \textit{absolutely fair} as defined in Definition~\ref{def:fair_loss}; \textbf{Benefit 2:} The loss of \textcolor{mismatch_color}{\textit{fp}} gets increased so that it can no longer deceive the network (detailed analysis in Sec.~\ref{subsec:analysis-of-auto-blur}). All values are calculated by OpenCV library, with Gaussian kernel size set to $4l+1$, $\sigma$ set to \textit{s.t.} $f_{gaussian}(0)=\frac{2}{3}, f_{gaussian}(l)=\frac{1}{6}, f_{gaussian}(2l)\approx 0$ and borders are zero-padding.
    }
    \label{fig:illustration_auto_blur}
\end{figure*}
Based on the OpenCV result of a specific case in Fig.~\ref{fig:illustration_auto_blur}, we give clear explanations on the effectiveness of the proposed Auto-Blur from two aspects.

\noindent\textbf{\colorbox[RGB]{255,200,233}{A Fair Judge.}}
As benefit 1 in Fig.~\ref{fig:illustration_auto_blur}e shows, being an \textit{absolutely fair loss} in a certain range as defined in Definition~\ref{def:fair_loss}, $\mathcal{L}_1 + \mathcal{L}_{ssim}$ photometric loss can fairly and accurately assess the network predictions.
While in baseline, no matter how much the predicted disparity deviates from \textit{gt}, loss remains unchanged.
The role of Auto-Blur is to `radiate' \textit{gt}'s characteristic color (red) to the surroundings (Fig.~\ref{fig:illustration_auto_blur}d left).
Moreover, this radiation is inversely proportional to the distance, so that as the distance between the predicted disparity and \textit{gt} gets smaller, the photometric information of \textit{gt} gets stronger, informing the network the current disparity is getting closer to \textit{gt}.
Thus, within a certain range, this strategy can make the penalty on the network increase proportionally with the degree of prediction error.

\noindent\textbf{\colorbox[RGB]{175,238,255}{Expose False-Positive.}}
When the assumption of photometric consistency breaks down, some false-positive (\textit{fp}) pixels may look more like the target pixel than \textit{gt}, thereby fooling the network.
(\textit{E.g.} in Fig.~\ref{fig:visualization}a, the pole and background's depths mixed together, since some white background pixels' color might incorrectly match the white pole).
As benefit 2 in Fig.~\ref{fig:illustration_auto_blur}e shows, Auto-Blur helps to increase penalty/loss of \textit{fp}, thus preventing it from matching the target pixel.
The benefit of Gaussian kernel's `color radiation' is to increase receptive fields when measuring pixel similarity, while keeping the original CNN kernel size.
Concretely, the \textit{fp} is radiated by surrounding blue (Fig.~\ref{fig:illustration_auto_blur}d right), whereas \textit{gt} is radiated by surrounding green (Fig.~\ref{fig:illustration_auto_blur}a bottom), thus making a difference.
Loss of \textit{fp} therefore gets increased, preventing it from deceiving the network.

\section{Conclusion}\label{sec:conclusion}

We examine the photometric loss of self-supervised MDE from the spatial frequency domain, and reveal two general issues rarely noticed by previous MDE researchers.
We draw a conclusion that the pixel-level ambiguity in the object junctions of input images is a more fundamental reason that hinders sharper depth edges.
Furthermore, we demonstrate that the photometric loss function cannot \textit{fairly} assess the predictions in high-freq regions, and could sometimes produce false global optimums.
Interestingly, we prove blurring the input images could reduce such \textit{unfairness} and efficiently enlarge receptive fields.
Our approaches are highly lightweight and versatile.
A large number of existing models get performance boosts from our methods, while no extra inference computation is needed at all.

\vspace{0.2cm}
\noindent\textbf{Acknowledgement.}
This work was supported by National Natural Science Foundation of China (No. 62172021).

{\small
\bibliographystyle{ieee_fullname}
\bibliography{main}
}

\end{document}